\documentclass{article}
\usepackage{aaai}
\usepackage{times}
\usepackage{helvet}
\usepackage{courier}
\usepackage{url}
\usepackage{enumerate}

\usepackage{graphicx}
\graphicspath{{./Images/}}

\usepackage{moreverb}
\usepackage{amsmath}
\usepackage{amssymb}
\usepackage{bm}
\def\cS{\mathcal{S}}
\def\cT{\mathcal{T}}
\def\cO{\mathcal{O}}
\def\cB{\mathcal{B}}
\def\cA{\mathcal{A}}

\renewcommand{\vec}[1]{\bm{#1}}
\def\vv{\vec{v}}

\usepackage[usenames]{color} 
\definecolor{Blue}{rgb}{0,0.16,0.90}
\definecolor{Red}{rgb}{0.90,0.16,0}
\definecolor{DarkBlue}{rgb}{0,0.08,0.45}
\definecolor{ChangedColor}{rgb}{0.9,0.08,0}

\usepackage[skip=2pt]{caption}          
\usepackage{mdwlist}

\title{Penetration Testing == POMDP Solving?}

\author{
  Carlos Sarraute\\
  Core Security Technologies \& ITBA\\
  Buenos Aires, Argentina\\
  \url{carlos@coresecurity.com}
  \And
  Olivier Buffet \and J\"org Hoffmann\\
  INRIA\\
  Nancy, France\\
  \url{{olivier.buffet,joerg.hoffmann}@loria.fr}
}

\nocopyright

\begin{document}
\maketitle

\begin{abstract}
Penetration Testing is a methodology for assessing network security,
by generating and executing possible attacks. Doing so automatically
allows for regular and systematic testing without a prohibitive amount
of human labor. A key question then is how to generate the
attacks. This is naturally formulated as a planning problem.  Previous
work \cite{LucSarRic10} used classical planning and hence ignores all
the incomplete knowledge that characterizes hacking. More recent work
\cite{SarRicLuc11} makes strong independence assumptions for the sake
of scaling, and lacks a clear formal concept of what the attack
planning problem actually {\em is}. Herein, we model that problem in
terms of partially observable Markov decision processes (POMDP). This
grounds penetration testing in a well-researched formalism,
highlighting important aspects of this problem's nature. 
POMDPs allow to model information gathering as an integral part of the
problem, thus providing for the first time a means to intelligently
mix scanning actions with actual exploits.

\end{abstract}

\section{Introduction}

Penetration Testing (short {\em pentesting}) is a methodology for
assessing network security, by generating and executing possible
attacks exploiting known vulnerabilities of operating systems and
applications (e.g., \cite{ArcGra04}). Doing so automatically allows
for regular and systematic testing without a prohibitive amount of
human labor, and makes pentesting more accessible to non-experts. A
key question then is how to automatically generate the attacks.

A natural way to address this issue is as an {\em attack planning}
problem. This is known in the AI Planning community as the ``Cyber
Security'' domain \cite{BodGohHaiHar05}. Independently (though
considerably later), the approach was put forward also by the
pentesting industry \cite{LucSarRic10}. The two domains essentially
differ only in the industrial context addressed. %
Herein, we are concerned exclusively with the specific context of
regular automatic pentesting, as in Core Security's ``Core Insight
Enterprise'' tool. We will use the term ``attack planning'' in that
sense.

Lucangeli et al.\ \shortcite{LucSarRic10} encoded attack planning into
PDDL, and used off-the-shelf planners. %
This already is useful,\footnote{In fact, this technology is currently
  employed in Core Security's commercial product, using a variant of
  Metric-FF.} however it is still quite limited. In particular, the
planning is classical---complete initial states and deterministic
actions---and thus not able to handle the uncertainty involved in this
form of attack planning. We herein contribute a planning model that
does capture this uncertainty, and allows to generate plans taking it
into account. To understand the added value of this technology, it is
necessary to examine the relevant context in some detail.

The pentesting tool has access to the details of the client
network. So why is there any uncertainty? The answer is simple:
pentesting is not Orwell's ``Big Brother''. Do {\em your} IT guys know
everything that goes on inside {\em your} computer?

It is safe to assume that the pentesting tool will be kept up-to-date
about the structure of the network, i.e., the set of machines and
their connections---these changes are infrequent and can easily be
registered. It is, however, impossible to be up-to-date regarding all
the details of the configuration of each machine, in the typical
setting where that configuration is ultimately in the hands of the
individual users. Thus, since the last series of attacks was
scheduled, the configurations may have changed, and the pentesting
tool does not know how exactly. Its task is to figure out whether any
of the changes open new dangerous vulnerabilities.

One might argue that the pentesting tool should first determine what
has changed, via {\em scanning} methods, and then address what is now
a classical planning problem involving only {\em exploits}, i.e.,
hacking actions modifying the system state. There are two flaws in
this reasoning: (a) scanning doesn't yield perfect knowledge so a
residual uncertainty remains; (b) scanning generates significant costs
in terms of running time and network traffic. So what we want is a
technique that (like a real hacker) can deal with uncertainty by
intelligently inserting scanning actions where they are useful for
scheduling the best exploits. To our knowledge, ours is the first work
that indeed offers such a method.

There is hardly any related work tackling uncertainty measures
(probabilities) in network security. The few works that exist (e.g.,
\cite{Bil03thesis,DawHal04}) are concerned with the defender's
viewpoint, and tackle a very different kind of uncertainty attempting
to model what an attacker would be likely to do. The above
mentioned work on classical planning is embedded into a pentesting
tool running a large set of scans as a pre-process, and afterwards
ignoring the residual uncertainty. This incurs both drawbacks (a) and
(b) above. The single work addressing (a) was performed in part by one
of the authors 
\cite{SarRicLuc11}. On the positive side, the proposed attack planner
demonstrates industrial-scale runtime performance, and in fact its
worst-case runtime is low-order polynomial. On the negative side, the
planner does not offer a solution to (b)---it still reasons only
about exploits, not scanning---and of course its efficiency is bought
at the cost of strong simplifying assumptions. Also, the work provides
no clear notion of what attack planning under uncertainty actually {\em
  is}.

Herein, we take the opposite extreme of the trade-off between accuracy
and performance. We tackle the problem in full, in particular
addressing information gathering as an integral part of the attack. We
achieve this by modeling the problem in terms of partially observable
Markov decision processes (POMDP). As a side effect, this modeling
activity serves to clarify some important aspects of this problem's
nature. A basic insight is that, whereas Sarraute et
al.\ \shortcite{SarRicLuc11} model the uncertainty as
non-deterministic actions---success probabilities of exploits---this
uncertainty is more naturally modeled as an uncertainty about {\em
  states}. The exploits as such are deterministic in that their
outcome is fully determined by the system
configuration.\footnote{Sometimes, non-deterministic effects are an
  adequate abstraction of state uncertainty, as in ``crossing the
  street''. The situation in pentesting is different because repeated
  executions will yield identical outcomes.}  Once this basic modeling
choice is made, all the rest falls into place naturally.

Our experiments are based on a problem generator that is not
industrial-scale realistic, but that allows to create reasonable test
instances by scaling the number of machines, the number of possible
exploits, and the time elapsed since the last activity of the
pentesting tool. Unsurprisingly, we find that POMDP solvers do not
scale to large networks. However, scaling is reasonable for individual
pairs of machines. As argued by Sarraute et
al.\ \shortcite{SarRicLuc11}, such pairwise strategies can serve as
the basic building blocks in a framework decomposing the overall
problem into two abstraction levels.

We next provide some additional background on pentesting and
POMDPs. We then detail our POMDP model of attack planning, and our
experimental findings. We close the paper with a brief discussion of
future work.

\section{Background}

We fill in some background on pentesting and POMDPs.

\subsection{Penetration Testing}

\begin{figure}[tbp]
  \centerline{
    \resizebox{0.85\linewidth}{!}{
      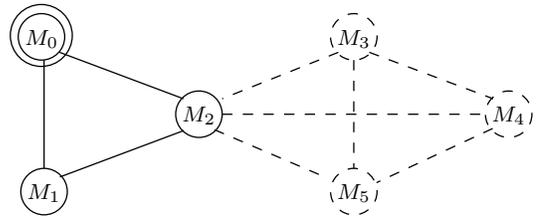
    }
  }
  \caption{An example network made of two sub-networks (cliques):
    $(M_0, M_1, M_2)$ (e.g., $M_0$ an outside computer, $M_1$ a web
    server and $M_2$ a firewall), and $(M_2, M_3, M_4, M_5)$. 1
    computer is under control ($M_0$), 2 are reachable ($M_1$ and
    $M_2$), and 3 are unreachable ($M_3$, $M_4$, $M_5$).}
  \label{fig:network}
\end{figure}

The objective of a typical penetration testing task is to gain control
over as many computers in a network as possible, with a preference for
some machines (e.g., because of their critical content). It starts
with one controlled computer: either outside the targeted network (so
that its first targets are machines accessible from the internet), or
inside this network (e.g., using a Trojan horse). As illustrated in
Figure~\ref{fig:network}, at any point in time one can distinguish
between 3 types of computers: those under control (on which an agent
has been installed, allowing to perform actions); those which are
reachable from a controlled computer because they share a sub-network
with one of them: and those which are unreachable from any controlled
computer.

Given currently controlled machines, one can perform two types of
actions targeting a reachable machine: tests---to identify its
configuration (OS, running applications, \dots)---, and exploits---to
install an agent by exploiting a vulnerability. A successful exploit
turns a reachable computer into a controlled one, and all its
previously unreachable neighbors into reachable computers.

A ``classic'' pentest methodology consists of a series of fixed steps,
for example:
\begin{itemize*}
\item perform a network discovery (obtain a list of all the reachable
  machines),
\item port scan all the reachable machines (given a fixed list of
  common ports, probe if they are open/closed/filtered),
\item given the previous information, perform OS detection module(s)
  on reachable machines (e.g., run nmap tests),
\item once the information gathering phase is completed, the following
  phase is to launch exploits against the (potentially vulnerable)
  machines.
\end{itemize*}
This could be improved---a long-term objective of this
work---as POMDP planning allows for more efficiency by mixing actions
from the different steps.

More details on pentesting will be given later when we describe how to
model it using the POMDP formalism.

\subsection{POMDPs}

POMDPs are usually defined \cite{Monahan82,Cassandra98thesis} by a tuple $\langle
\cS, \cA, \cO, T, O, r, b_0 \rangle$ where, at any time step,
the system being in some state $s \in \cS$ (the {\em state space}),
the agent performs an action $a\in \cA$ (the {\em action space}) that
results in (1) a transition to a state $s'$ according to the {\em
  transition function} $T(s,a,s')=Pr(s'| s,a)$, (2) an observation
$o\in\cO$ (the {\em observation space}) according to the {\em
  observation function} $O(s',a,o)=Pr(o| s',a)$ and (3) a scalar {\em
  reward} $r(s,a)$. $b_0$ is the initial probability distribution over
states. Unless stated otherwise, the sets $\cS$, $\cA$ and $\cO$ are finite.

In this setting, the problem is for the agent to find a decision {\em
  policy} $\pi$ choosing, at each time step, the best action based on
its past observations and actions so as to maximize its future gain
(which can be measured for example through the total accumulated
reward). Compared to classical
deterministic planning, the agent has to face the difficulty in
accounting for a system not only with uncertain dynamics but also whose
current state is imperfectly known.

The agent typically reasons about the hidden state of the system using
a {\em belief state} $b \in \cB=\Pi(\cS)$ (the set of probability
distributions over $\cS$) using the following Bayesian update formula
when performing action $a$ and observing $o$:
\begin{align*}
  b^{a,o}(s') &= \frac{O(s',a,o)}{Pr(o| a,b)}\sum_{s\in \cS} T(s,a,s') b(s),
\end{align*}
where $Pr(o| a,b)=\sum_{s,s''\in \cS}O(s'',a,o) T(s,a,s'')
b(s)$. Using belief states, a POMDP can be rewritten as an MDP over
the belief space, or {\em belief MDP}, $\langle \cB, \cA, \cT, \rho
\rangle$, where the new transition and reward functions are both
defined over $\cB\times\cA\times\cB$. With this reformulation, a
number of theoretical results about MDPs can be extended, such as the
existence of a deterministic policy that is optimal. An issue is that this
belief MDP is defined over a continuous---and thus infinite---belief
space.

For a finite horizon\footnote{In practice we consider an infinite
  horizon.} $T>0$ the objective is to find a policy verifying $\pi^* =
\arg\max_{\pi \in \cA^{\cB}} J^\pi(b_0)$ with
\begin{align*}
  J^{\pi}(b_0) &= E\left[ \sum_{t=0}^{T-1} \gamma^t r_t \middle| b_0, \pi \right],
\end{align*}
where $b_0$ is the initial belief state, $r_t$ the reward obtained at
time step $t$, and $\gamma \in (0,1)$ a discount factor. Bellman's principle of
optimality~\cite{Bellman54dynprog} lets us compute this function
recursively through the {\em value function}
\begin{align*}
  V_n(b) & = \max_{a\in\cA} \left[ \rho(b,a) + \beta \sum_{b' \in
      \cB} \phi(b,a,b')V_{n-1}(b') \right],
\end{align*}
where, for all $b\in \cB$, $V_0(b)=0$, and $J^{\pi}(b)=V_{n=T}(b)$.

For our experiments we use SARSOP \cite{KurHsuLee08}, a state of the
art point-based algorithm, i.e., an algorithm approximating the value
function as the upper envelope of a set of hyperplanes, these
hyperplanes corresponding to a selection of particular belief points.

\section{Modeling Penetration Testing with POMDPs}

As penetration testing is about acting under partial
observability, POMDPs are a natural candidate to model this particular
problem. They allow to model the problem of knowledge acquisition and
to account for probabilistic information, e.g., the fact that certain
configurations or vulnerabilities are more frequent than others. In
comparison, classical planning approaches \cite{LucSarRic10} assume
that the whole network configuration is known, so that no exploration
is required.
The present section discusses how to formalize penetration testing
using POMDPs. As we shall see, the uncertainty is located essentially
in the initial belief state.
This is different from modeling the uncertainty in pentesting using
probabilistic action outcomes as in \cite{SarRicLuc11}, which does not
account for the real dynamics of the system. Also, as indicated
previously, unlike our POMDPs, the approach of Sarraute et
al.\ \shortcite{SarRicLuc11} only chooses exploits, assuming
a naive {\em a priori} knowledge acquisition and thus ignoring the
interaction between these two.

\subsection{States}

First, any sensible penetration test will have a finite
execution. There is nothing to be gained here by infinitely executing
a looping behavior. Every pentest terminates either when some event
(e.g., an attack detection) stops it, or when the additional access
rights that could yet be gained (from the finite number of access
rights) do not outweigh the associated costs. %
This implies that there exists an absorbing {\em terminal} state and
that we are solving a Stochastic Shortest Path problem (SSP).

Then, in the context of pentesting, we do not need the full state of the
system to describe the current situation. We will thus focus on
aspects that are relevant for the task at hand.
This state for example does not need to comprise the network topology
as it is assumed here to be static and known. But it will have to
account for the configuration and status of each computer on the
network.

A computer's {\em configuration} needs to describe the applications
present on the computer and that may (i) be vulnerable or (ii) reveal
information about potentially vulnerable applications. This comprises
its operating system (OS) as well as server applications for the web,
databases, email, ... The description of an application does not need
to give precise version numbers, but should give enough details to
know which (known) vulnerabilities are present, or what information
can be obtained about the system. For example, the open ports on a
given computer are aspects of the OS that may reveal not only the OS
but also which applications it is running.

The computers' configurations (and the network topology) give a {\em
  static} picture of the system independently of the progress of the
pentest. To account for the current situation one needs to specify,
for each computer, whether a given agent has been installed on it,
whether some applications have crashed (e.g., due to the failure of an
exploit), and which computers are accessible. Which computers are
accessible depends only on the network topology and on where agents
have been installed, so that there is no need to explicitly add this
information in the state.
Fig.~\ref{fig:States-M0} gives a {\tt states} section from an actual
POMDP file (using the file format of Cassandra's toolbox) in a setting
with a single machine {\tt M0}, which is always accessible (not
mentioning the computer from which the pentest is started).

\begin{figure}[tbp]
  {
    \scriptsize
    \begin{minipage}{0.48\linewidth}
      \begin{verbatimtab}
states :
terminal
M0-win2000
M0-win2000-p445
M0-win2000-p445-SMB
M0-win2000-p445-SMB-vuln
M0-win2000-p445-SMB-agent
M0-win2003
M0-win2003-p445
M0-win2003-p445-SMB
\end{verbatimtab}
\end{minipage}
\hfill
\begin{minipage}{0.48\linewidth}
\begin{verbatimtab}
M0-win2003-p445-SMB-vuln
M0-win2003-p445-SMB-agent
M0-winXPsp2
M0-winXPsp2-p445
M0-winXPsp2-p445-SMB
M0-winXPsp2-p445-SMB-vuln
M0-winXPsp2-p445-SMB-agent
M0-winXPsp3
M0-winXPsp3-p445
M0-winXPsp3-p445-SMB
      \end{verbatimtab}
    \end{minipage}
  }
  \caption{A list of states in a setting with a single computer (M0)
    which can be a Windows 2000, 2003, XPsp2 or XPsp3, may have port
    445 open and, if so, may be running a SAMBA server which may be
    vulnerable (except for XPsp3) and whose vulnerability may have
    been exploited.}
  \label{fig:States-M0}
\end{figure}

Note that a computer's configuration should also provide information
on whether having access to it is valuable in itself, e.g., if there
is valuable data on its hard drive. This will be used when defining
the reward function.

\subsection{Actions (\& Observations)}

First, we need a {\tt Terminate} action that can be used to reach the
{\tt terminal} state voluntarily. Note that specific outcomes of
certain actions could also lead to that state.

Because we assume that the network topology is known {\em a priori},
there is no need for actions to discover reachable machines. We are
thus left with two types of actions: {\em tests}, which allow to
acquire information about a computer's configuration, and {\em
  exploits}, which attempt to install an agent on a computer by
exploiting a vulnerability. Fig.~\ref{fig:Actions-M0} lists actions in
our running example started in Fig.~\ref{fig:States-M0}.

\begin{figure}[tbp]
  {
    \scriptsize
    \begin{minipage}{0.48\linewidth}
      \begin{verbatimtab}
actions :
Terminate
Probe-M0-p445
OSDetect-M0
\end{verbatimtab}
\end{minipage}
\hfill
\begin{minipage}{0.48\linewidth}
\begin{verbatimtab}

Exploit-M0-win2000-SMB
Exploit-M0-win2003-SMB
Exploit-M0-winXPsp2-SMB
      \end{verbatimtab}
    \end{minipage}
  }
  \caption{A list of actions in the same setting as
    Fig.~\ref{fig:States-M0}, with 1 {\tt Terminate} action, 2 tests,
    and 3 possible exploits.}
  \label{fig:Actions-M0}
\end{figure}

\subsubsection{Tests}

Tests are typically performed using programs such as {\em nmap}
\cite{Fyodor98}, which scans a specific computer for open ports and,
by analyzing the response behavior of ports, allows to make guesses
about which OS and services are running. Note that such observation
actions have a cost either in terms of time spent performing analyses,
or because of the probability of being detected due to the generated
network activity. This is the reason why one has to decide which
tests to perform rather than perform them all.

In our setting, we only consider two types of tests:
\begin{description*}
\item[OS detection:] A typical OS detection will return a list of
  possible OSes, the ones likely to explain the observations of the
  analysis tool. As a result, one can prune from the belief state
  (=set to zero probability) all the states corresponding with
  non-matching OSes, and then re-normalize the remaining non-zero
  probabilities.

  Keeping with the same running example,
  Fig.~\ref{fig:OSDetect-M0} presents the transition and observation
  models associated with action {\tt OSDetect-M0}, which can
  distinguish winXP configurations from win2000/2003; and following is
  an example of the evolution of the belief state:
  \\
  \centerline{\scriptsize
    \begin{tabular}{c@{ $($}c@{,}c@{,}c@{,}c@{,}c@{,}c@{,}c@{,}c@{,}c@{,}c@{,}c@{,}c@{,}c@{,}c@{,}c@{,}c@{,}c@{,}c@{,}c@{$)$}}
      initial
      & 0& 0& 0& 0& 0& 0& $\frac{1}{8}$& $\frac{1}{8}$& $\frac{1}{8}$& $\frac{1}{8}$& 0& $\frac{1}{8}$& $\frac{1}{8}$& $\frac{1}{8}$& $\frac{1}{8}$& 0& 0& 0& 0 \\ \hline
      {\tt winXP}
      & 0& 0& 0& 0& 0& 0& 0& 0& 0& 0& 0& $\frac{1}{4}$& $\frac{1}{4}$& $\frac{1}{4}$& $\frac{1}{4}$& 0& 0& 0& 0 \\
      {\tt win2000/2003}
      & 0& 0& 0& 0& 0& 0& $\frac{1}{4}$& $\frac{1}{4}$& $\frac{1}{4}$& $\frac{1}{4}$& 0& 0& 0& 0& 0& 0& 0& 0& 0
    \end{tabular}
  }
  
  \begin{figure}[tbp]
    {
      \scriptsize
      \begin{verbatimtab}
T: OSDetect-M0 identity
O: OSDetect-M0: * : * 0
O: OSDetect-M0: * : undetected 1
O: OSDetect-M0: M0-win2000               : win 1
O: OSDetect-M0: M0-win2000-p445          : win 1
O: OSDetect-M0: M0-win2000-p445-SMB      : win 1
O: OSDetect-M0: M0-win2000-p445-SMB-vuln : win 1
O: OSDetect-M0: M0-win2000-p445-SMB-agent: win 1
O: OSDetect-M0: M0-win2003               : win 1
O: OSDetect-M0: M0-win2003-p445          : win 1
O: OSDetect-M0: M0-win2003-p445-SMB      : win 1
O: OSDetect-M0: M0-win2003-p445-SMB-vuln : win 1
O: OSDetect-M0: M0-win2003-p445-SMB-agent: win 1
O: OSDetect-M0: M0-winXPsp2               : winxp 1
O: OSDetect-M0: M0-winXPsp2-p445          : winxp 1
O: OSDetect-M0: M0-winXPsp2-p445-SMB      : winxp 1
O: OSDetect-M0: M0-winXPsp2-p445-SMB-vuln : winxp 1
O: OSDetect-M0: M0-winXPsp2-p445-SMB-agent: winxp 1
O: OSDetect-M0: M0-winXPsp3               : winxp 1
O: OSDetect-M0: M0-winXPsp3-p445          : winxp 1
O: OSDetect-M0: M0-winXPsp3-p445-SMB      : winxp 1
\end{verbatimtab}
    }
    \caption{Transition and observation models for action {\tt
        OSDetect-M0}. The first line specifies that this action's
      transition matrix is the identity matrix. The remaining lines
      describe this action's observation function by giving the
      probability (here 0 or 1) of each possible state-observation
      pair, defaulting to the {\tt undetected} observation for all
      states.}
    \label{fig:OSDetect-M0}
  \end{figure}
\item[Port scan:] Scanning port $X$ simply tells if it is
  open or closed; by pruning from the belief state the states that
  match the open/closed state of port $X$, one implicitely
  refines which OS and applications may be running.

  Action {\tt Probe-M0-p445}, for example, is modeled as depicted on
  Fig.~\ref{fig:Probe-M0-p445} and could give the following evolution:
  \\
  \centerline{\scriptsize
    \begin{tabular}{c@{ $($}c@{,}c@{,}c@{,}c@{,}c@{,}c@{,}c@{,}c@{,}c@{,}c@{,}c@{,}c@{,}c@{,}c@{,}c@{,}c@{,}c@{,}c@{,}c@{$)$}}
      initial
      & 0& 0& 0& 0& 0& 0& $\frac{1}{8}$& $\frac{1}{8}$& $\frac{1}{8}$& $\frac{1}{8}$& 0& $\frac{1}{8}$& $\frac{1}{8}$& $\frac{1}{8}$& $\frac{1}{8}$& 0& 0& 0& 0 \\ \hline
      {\tt open-port}
      & 0& 0& 0& 0& 0& 0& 0& $\frac{1}{6}$& $\frac{1}{6}$& $\frac{1}{6}$& 0& 0& $\frac{1}{6}$& $\frac{1}{6}$& $\frac{1}{6}$& 0& 0& 0& 0 \\
      {\tt closed-port}
      & 0& 0& 0& 0& 0& 0& $\frac{1}{2}$& 0& 0& 0& 0& $\frac{1}{2}$& 0& 0& 0& 0& 0& 0& 0
    \end{tabular}
  }

  \begin{figure}[tbp]
    {
      \scriptsize
      \begin{verbatimtab}
T: Probe-M0-p445 identity
O: Probe-M0-p445: * : * 0
O: Probe-M0-p445: * : closed-port 1
O: Probe-M0-p445: M0-win2000-p445           : open-port 1
O: Probe-M0-p445: M0-win2000-p445-SMB       : open-port 1
O: Probe-M0-p445: M0-win2000-p445-SMB-vuln  : open-port 1
O: Probe-M0-p445: M0-win2000-p445-SMB-agent : open-port 1
O: Probe-M0-p445: M0-win2003-p445           : open-port 1
O: Probe-M0-p445: M0-win2003-p445-SMB       : open-port 1
O: Probe-M0-p445: M0-win2003-p445-SMB-vuln  : open-port 1
O: Probe-M0-p445: M0-win2003-p445-SMB-agent : open-port 1
O: Probe-M0-p445: M0-winXPsp2-p445          : open-port 1
O: Probe-M0-p445: M0-winXPsp2-p445-SMB      : open-port 1
O: Probe-M0-p445: M0-winXPsp2-p445-SMB-vuln : open-port 1
O: Probe-M0-p445: M0-winXPsp2-p445-SMB-agent: open-port 1
O: Probe-M0-p445: M0-winXPsp3-p445          : open-port 1
O: Probe-M0-p445: M0-winXPsp3-p445-SMB      : open-port 1
      \end{verbatimtab}
    }
    \caption{Transition and observation models for action {\tt
        Probe-M0-p445}. The transition is again the identity, and the
      observation is {\tt closed-port} by default, and {\tt open-port}
      for all states in which port 445 is open.}
    \label{fig:Probe-M0-p445}
  \end{figure}
\end{description*}

Note that a test has no state outcome (the state remains the same),
and that its observation outcome is considered as deterministic: given
the---real, but hidden---configuration of a computer, a given test
always returns the same observation.
Another interesting point is that (i) tests provide information about
computer configurations and (ii) computer configurations are static,
so that there is no use repeating a test as it cannot provide or
update any information.

\subsubsection{Exploits}

Exploits make use of an application's vulnerability to gain (i) some
control over a computer from another computer (remote exploit), or
(ii) more control over a computer (local exploit / privilege
escalation). Local exploits do not differ significantly from remote
exploits since it amounts to considering each privilege level as
a different (virtual) computer in a sub-network. As a consequence, for
the sake of clarity, we only consider one privilege level per
computer.

More precisely, we consider that any successful exploit will provide
the same control over the target computer, whatever the exploit and
whatever its configuration. This allows (i) to assume that the same
set of actions is available on any controlled computer, and (ii) to
avoid giving details about which type of agent is installed on a
computer.

The success of a given exploit action $E$ depends deterministically on
the configuration of the target computer, so that: (i) there is no use
in attempting an exploit $E$ if none of the probable configurations is
compatible with this exploit, and (ii) the outcome of $E$---either
success or failure---provides information about the configuration of
the target. In the present paper, we even assume that a computer's
configuration is completely observed once it is under control.

{\tt Exploit-M0-win2003-SMB} is modeled in
Fig.~\ref{fig:Exploit-M0-win2003-SMB}, and an example evolution of the
belief under this action is:
\\
\centerline{\scriptsize
    \begin{tabular}{c@{ $($}c@{,}c@{,}c@{,}c@{,}c@{,}c@{,}c@{,}c@{,}c@{,}c@{,}c@{,}c@{,}c@{,}c@{,}c@{,}c@{,}c@{,}c@{,}c@{$)$}}
      initial
      & 0& 0& 0& 0& 0& 0& $\frac{1}{8}$& $\frac{1}{8}$& $\frac{1}{8}$& $\frac{1}{8}$& 0& $\frac{1}{8}$& $\frac{1}{8}$& $\frac{1}{8}$& $\frac{1}{8}$& 0& 0& 0& 0 \\ \hline
      {\tt success}
      & 0& 0& 0& 0& 0& 0& 0& 0& 0& 0& 0& 0& 0& 0& 0& 1& 0& 0& 0 \\
      {\tt failure}
      & 0& 0& 0& 0& 0& 0& $\frac{1}{7}$& $\frac{1}{7}$& $\frac{1}{7}$& $\frac{1}{7}$& 0& $\frac{1}{7}$& $\frac{1}{7}$& $\frac{1}{7}$& 0& 0& 0& 0& 0
    \end{tabular}
  }

  \begin{figure}[tbp]
    {
      \scriptsize
      \begin{verbatimtab}
T: Exploit-M0-win2003-SMB identity
T: Exploit-M0-win2003-SMB: M0-win2003-p445-SMB-vuln
                             : * 0
T: Exploit-M0-win2003-SMB: M0-win2003-p445-SMB-vuln
                             : M0-win2003-p445-SMB-agent 1
O: Exploit-M0-win2003-SMB: * : * 0
O: Exploit-M0-win2003-SMB: * : no-agent 1
O: Exploit-M0-win2003-SMB: M0-win2003-p445-SMB-agent
                             : agent-installed 1
      \end{verbatimtab}
    }
    \caption{Transition and observation models for action {\tt
        Exploit-M0-win2003-SMB}. The transition is the identity except
      if M0 is vulnerable, where an agent gets installed. The
      observation is {\tt no-agent} by default and {\tt
        agent-installed} if the exploit is successful.}
    \label{fig:Exploit-M0-win2003-SMB}
  \end{figure}

\subsection{Rewards}

First, no reward is received when the {\tt Terminate} action is used,
or once the terminal state is reached.
Otherwise, the reward function has to account for various things:
\begin{description*}
\item[Value of a computer ($r_c$):] The objective of a pentest is to
  gain access to a number of computers. Here we thus propose to assign
  a fixed reward for each successful exploit (on a previously
  uncontrolled machine). In a more realistic setting, one could reward
  accessing for the first time a given valuable data, whatever
  computer hosts these data.
\item[Time is money ($r_t$):] Each action---may it be a test or an
  exploit---has a duration, so that the expected duration of the pentest
  may be minimized by assigning each transition a cost (negative
  reward) proportional to its duration. One could also
  consider a maximum time for the pentest rather than minimizing it.
\item[Risk of detection ($r_d$):] We do not explicitely model the
  event of being detected (that would lead to the terminal state with
  an important cost), but simply consider transition costs that depend
  on the probability of being detected.
\end{description*}
As a result, a transition $s,a,s'$ comes with a reward that is the sum
of these three components: $r=r_c+r_t+r_d$. Although some rewards are
positive, we are still solving an SSP since such positive rewards
cannot be received multiple times and thus cyclic behavior is not
sensible.

\subsection{POMDP Model Generation}

Generating a POMDP model for pentesting requires knowledge about
possible states, actions, and observations, plus the reward function
and the initial belief state.  Note first that the POMDP model may
evolve from one pentest to the next due to new applications, exploits
or tests.

Action and observation models for the various possible tests and
exploits can be derived from the documentation of testing tools (see,
e.g., nmap's manpage) and databases such as CVE (Common
Vulnerabilities and
Exposures)\footnote{http://cve.mitre.org/}. Information could
presumably be automatically extracted from such databases, which are
already very structured. In our experiments, we start from a
proprietary database of Core Security Technologies. %
The two remaining components of the model---the reward function and
the initial belief state---involve quantitative information which is
more difficult to acquire. In our experiments, this information is
estimated based on expert knowledge. 

Regarding rewards, statistical models can be used to estimate, for any
particular action, the probability of being detected, and the
probabilistic model of its duration. But a human decision is required
to assign a value for the cost of a detection, for gaining control
over one target computer or the other, and for spending a certain
amount of time.


The definition of the initial belief state is linked to the fact that
penetration testing is a task repeated regularly, and has access to
previous pentesting reports on the same network. The pentester thus
has knowledge about the previous configuration of the network
(topology and machines), and which weaknesses have been reported. This
information, plus knowledge of typical update behaviors (applying
patches or not, downloading service packs...), allows an informed
guess on the current configuration of the network.

\begin{figure}[tbp]
  \centerline{
    \resizebox{0.88\linewidth}{!}{
      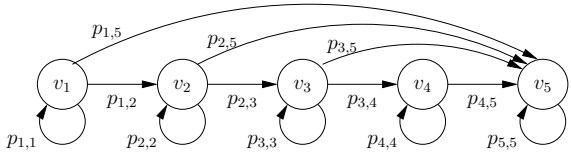
    }
  }
  \caption{Markov Chain modeling an application's updating
    process. Vertices are marked with version numbers, and edges with
    transition probabilities per time step.}
  \label{fig:updating-MC}
\end{figure}

We propose to mimick this reasoning to compute the initial belief
state. To keep things simple, we only consider a basic software update
behavior (assuming that softwares are independent from each other):
each day, an application may probabilistically stay unchanged, or be
upgraded to the next version or to the latest version. The updating
process of a given application can then be viewed as a Markov chain as
illustrated in Fig.~\ref{fig:updating-MC}. Assuming that (i) the
belief about a given application version was, at the end of the last
pentest, some vector $\vv_0$, and (ii) $T$ days (the time unit in the
Markov chain) have passed, then this belief will have to be updated as
$\vv_T = U^T \vv_0$, where $U$ is the matrix representation of the
chain. For Fig.~\ref{fig:updating-MC}, this matrix reads:
\begin{align*}
  U &=
  \left(
    \begin{array}{ccccc}
      p_{1,1} & 0 & 0 & 0 & 0 \\
      p_{1,2} & p_{2,2} & 0 & 0 & 0 \\
      0 & p_{2,3} & p_{3,3} & 0 & 0 \\
      0 & 0 & p_{3,4} & p_{4,4} & 0 \\
      p_{1,5} & p_{2,5} & p_{3,5} & p_{4,5} & p_{5,5} \\
    \end{array}
  \right).
\end{align*}

\noindent
This provides a factored approach to compute initial belief states. Of
course, in this form the approach is very simplistic. A realistic
method would involve elaborating a realistic model of system
development. This is a research direction in its own right. We come
back to this at the end of the paper.

\begin{figure*}[t]
\begin{minipage}{.45\textwidth}
  \centering
  (a) \includegraphics[width=0.85\textwidth]{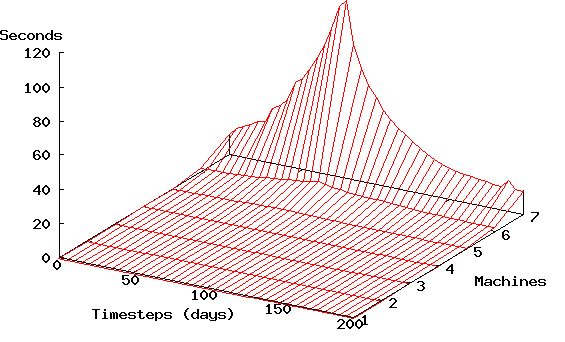}
\end{minipage}
\hfill
\begin{minipage}{.45\textwidth}
  \centering
  (b) \includegraphics[width=0.85\textwidth]{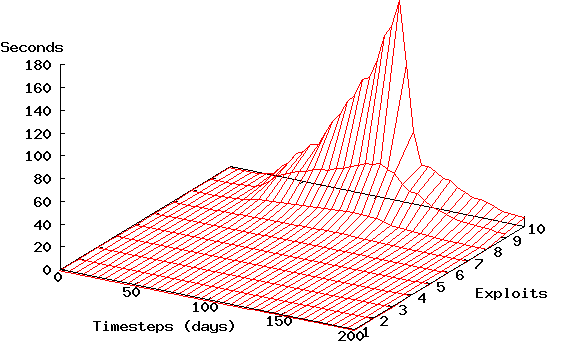}
\end{minipage}

\begin{minipage}{.45\textwidth}
  \centering
  (c) \includegraphics[width=0.85\textwidth]{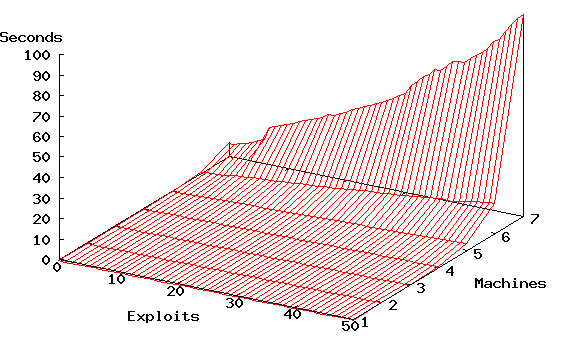}
\end{minipage}
\hfill
\begin{minipage}{.45\textwidth}
  \centering
  (d) \includegraphics[width=0.85\textwidth]{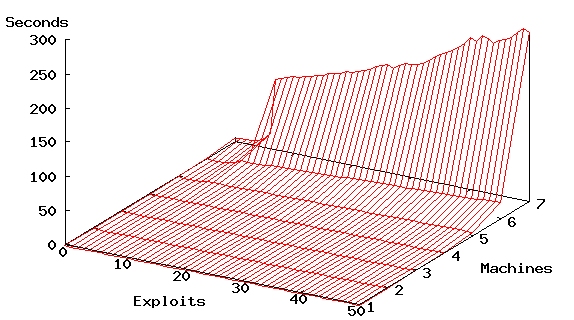}
\end{minipage}
  \caption{\label{fig:combined-scaling} POMDP solver runtime ($z$
    axis) when scaling: (a) time delay vs.\ the number of machines,
    (b) time delay vs.\ the number of exploits, (c) machines
    vs.\ exploits with time delay 10, and (d) machines vs.\ exploits
    with time delay 80.}
\vspace{-0.0cm}
\end{figure*}

\section{Solving Penetration Testing with POMDPs}

We now describe our experiments. We first fill in some details on the
setup, then discuss different scaling scenarios, before having a
closer look at some example policies generated by the POMDP solver.

\subsection{Experiments Setup}

The experiments are run on a machine with an Intel Core2 Duo CPU at
2.2 GHz and 3 GB of RAM. We use the APPL (Approximate POMDP Planning)
toolkit\footnote{APPL 0.93 at
  http://bigbird.comp.nus.edu.sg/pmwiki/farm/appl/}.  This C++
implementation of the SARSOP algorithm is easy to compile and use, and
has reasonable performance. The solver is run without time horizon
limit, until a target precision $\epsilon = 0.001$ is reached. Since
we are solving a stochastic shortest path problem, a discount factor
is not required, however we use $\gamma = 0.95$ to improve
performance. We will briefly discuss below the effect of changing
$\epsilon$ and $\gamma$.

Our problem generator is implemented in Python. It has 3 parameters:
\begin{itemize*}
\item number of machines $M$ in the target network,
\item number of exploits $E$ in the pentesting tool, that are
  applicable in the target network,
\item time delay $T$ since the last pentest, measured in days.
\end{itemize*}
For simplicity we assume that, at time $T = 0$, the information about
the network is perfect, i.e., there is no uncertainty. As $T$ grows,
uncertainty increases as described in the previous section, where the
parameters of the underlying model, cf.\ Fig.~\ref{fig:updating-MC},
are estimated by hand. The network topology consists of $1$ outside
machine and $M-1$ other machines in a fully connected network. The
configuration details are scaled along with $E$, i.e., details are
added as relevant for the exploits (note that irrelevant configuration
details would not serve any purpose in this application). As
indicated, the exploits are taken from a Core Security database which
contains the supported systems for each exploit (specific OS and
application versions that are vulnerable). The $E$ exploits are
distributed evenly over the $M$ machines. We require that $E \geq M$
so that each machine gets at least one exploit (otherwise the machine
could be removed from the encoding).

\subsection{Combined Scaling}

We discuss performance---solver runtime---as a function of $M$, $E$,
and $T$. To make data presentation feasible, at any one time we scale
only 2 of the parameters.

Consider first Figure~\ref{fig:combined-scaling} (a), which scales $M$
and $T$. $E$ is fixed to the minimum value, i.e., each machine has a
fixed OS version and one target application. In this setting, there
are $3^M$ states. For $M = 8$, the generated POMDP file has $6562$
states and occupies $71$ MB on disk; the APPL solver runs out of
memory when attempting to parse it. Thus, in this and all experiments
to follow, $M \leq 7$.

Naturally, runtime grows exponentially with $M$---after all, even the
solver input does. As for $T$, interestingly this exhibits a very
pronounced easy-hard-easy pattern. Investigating the reasons for this,
we found that it is due to a low-high-low pattern of the ``amount of
uncertainty'' as a function of $T$. Intuitively, as $T$ increases, the
probability distribution in the initial belief state first becomes
``broader'' because more application updates are possible. Then, after
a certain point, the probability mass accumulates more and more ``at
the end'', i.e., at the latest application versions, and the
uncertainty decreases again. Formally, this can be captured in terms
of the entropy of $b_0$, which exhibits a low-high-low pattern
reflecting that of Figure~\ref{fig:combined-scaling} (a).

In Figure~\ref{fig:combined-scaling} (b), scaling the number $E$ of
exploits as well as $T$, the number of machines is fixed to 2 (the
localhost of the pentester, and one target machine. We observe the same
easy-hard-easy pattern over $T$. As with $M$, runtime grows
exponentially with $E$ (and must do so since the solver input
does). However, with small or large $T$, the exponential behavior does
not kick in until the maximum number of exploits, 10, that we consider
here. This is important for practice since small values of $T$ (up to
$T=50$) are rather realistic in regular pentesting. In the next
sub-section, we will examine this in more detail to see how far we can
scale $E$, in the 2-machines case, with small $T$.

Figure~\ref{fig:combined-scaling} (c) and (d) show the combined
scaling over machines and exploits, for a favorable value of $T$
($T=10$, (c)) and an unfavorable one ($T=80$, (d)). Here the behavior
is rather regular. By all appearances, it grows exponentially in both
parameters. An interesting observation is that, in (c), the growth in
$M$ kicks in earlier, and rather more steeply, than in (d). 
Note that, in (d), the curve over $E$ flattens around
$T=10$. We discuss this behavior in the next sub-section.

To give an impression on the effect of the discount factor on solver
performance, with $M=2, E=11, T=40$, solver
runtime goes from 17.77 s (with $\gamma = 0.95$) to 279.65 s (with
$\gamma = 0.99$). APPL explicitly checks that $\gamma < 1$, so $\gamma
= 1$ could not be tried.  With our choice $\gamma = 0.95$ we still get
good policies (cf. further below).

\subsection{The 2-Machines Case}

\begin{figure}[htb]
\vspace{-0.25cm}
\centerline{\includegraphics[width=0.45\textwidth]{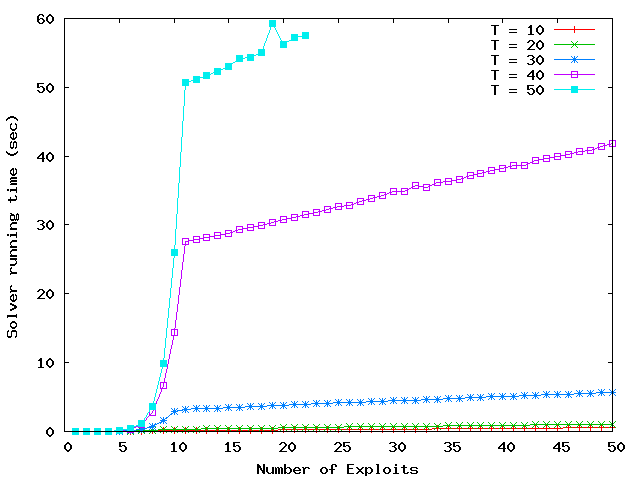}}
\vspace{-0.1cm}
  \caption{\label{fig:2-machines} POMDP solver runtime when scaling
    the number of exploits, for different realistically small settings
    of the time delay, in the 2-machines case.}
\vspace{-0.25cm}
\end{figure}

As hinted, the 2-machines case is relevant because it may serve as the
``atomic building block'' in an industrial-scale solution, cf.\ also
the discussion in the outlook below. The question then is whether or
not we can scale the number of exploits into a realistic region. We
have seen above already that this is not possible for unfavorable
values of $T$. However, are these values to be expected in practice?
As far as Core Security's ``Core Insight Enterprise'' tool goes, the
answer is ``no''. In security aware environments, pentesting should be
performed at regular intervals of at most 1 month. Consequently,
Figure~\ref{fig:2-machines} shows data for $T \leq 50$.

For the larger values of $T$, the data shows a very steep incline
between $E=5$ and $E=10$, followed by what appears to be linear
growth. This behavior is caused by an unwanted bias in our current
generator.\footnote{%
The exploits to be added are ordered in a way so that their likelihood
of succeeding decreases monotonically with $|E|$. After a certain
point, they are too unlikely to affect the policy quality by more than
the target precision $\epsilon$. The POMDP solver appears to determine
this effectively. } Ignoring this phenomenon, what matters to us here
is that, for the most realistic values of $T$ ($T=10,20$), scaling is
very good indeed, showing no sign of hitting a barrier even at
$E=50$.
Of course, this result must be qualified against the realism of the
current generator. %
It remains an open question whether similar scaling will be achieved
for more realistic simulations of network development.

\subsection{POMDPs make Better Hackers}

As an illustration of the policies found by the POMDP solver, consider
a simple example wherein the pentester has 4 exploits: an SSH exploit
(on OpenBSD, port 22), a wu-ftpd exploit (on Linux, port 21), an IIS
exploit (on Windows, port 80), and an Apache exploit (on Linux, port
80). The probability of the target machine being Windows is higher than
the probability of the other OSes.

Previous automated pentesting methods, e.g.\ Lucangeli {\em et
  al.}\ \shortcite{LucSarRic10}, proceed by first performing a port
scan on common ports, then executing OS detection module(s), and
finally launching exploits for potentially vulnerable services.

With our POMDP model, the policy obtained is to first test whether
port 80 is open, because the expected reward is greater for the two
exploits which target port 80, than for each of the exploits for port
21 or 22.  If port 80 is open, the next action is to launch the IIS
exploit for port 80, skipping the OS detection because Windows is more
probable than Linux, and the additional information that OS Detect can
provide doesn't justify its cost (additional running time). %
If the exploit is successful, terminate. Otherwise, continue with the
Apache exploit (not probing port 80 since that was already done), and
if that fails then probe port 21, etc.

In summary, the policy orders exploits by promise, and executes port
probe and OS detection actions on demand where they are
cost-effective. %
This improves on Sarraute et al.\ \shortcite{SarRicLuc11}, whose
technique is capable only of ordering exploits by promise. %
What's more, practical cases typically involve exploits whose outcome
delivers information about the success probability of other exploits,
due to common reasons for failure---exploitation prevention
techniques. Then the best ordering
of exploits depends on previous exploits' outcome. POMDP policies
handle this naturally, however it is well beyond the capabilities of
Sarraute et al.'s approach. We omit the details for space reasons.

\section{Discussion}

POMDPs can model pentesting more naturally and accurately than
previously proposed planning-based models
\cite{LucSarRic10,SarRicLuc11}. While, in general, scaling is limited,
we have seen that it appears reasonable in the 2-machines case where
we are considering only how to get from one machine to another. An
idea to use POMDP reasoning in practice is thus to perform it for all
connected pairs of machines in the network, and thereafter use these
solutions as the input for a high-level planning procedure. That
procedure would consider the pairwise solutions to be atomic, i.e., no
backtracking over these decisions would be made. Indeed, this is one
of the abstractions made---successfully, as far as runtime performance
is concerned---by Sarraute et al.\ \shortcite{SarRicLuc11}. Our
immediate future work will be to explore whether a POMDP-based
solution of this type is useful, the question being how large the
overhead for planning all pairs is, and how much of the solution
quality gets retained at the global level.

A line of basic research highlighted by our work is the exploitation
of special structures in POMDPs. First, in our model, all actions are
deterministic. Second, some of the uncertain parts of the state (e.g.\
the operating systems) are static, for the purpose of pentesting, in
the sense that none of the actions affect them. Third, unless one
models possible detrimental side-effects of exploits (cf.\ directly
below), pentesting is ``monotonic'': accessibility, and thus the set
of actions applicable, can only grow. Fourth, any optimal policy will
apply each action at most once. Finally, some aspects of the
state---in particular, which computers are controlled and
reachable---are directly visible and could be separately modeled as
being such. To our knowledge, this last property alone has been
exploited in POMDP solvers (e.g.,\ \cite{AraThoBufCha-ICTAI10}), and the only
other property mentioned in the literature appears to be the first one
(e.g.,\ \cite{Bon09}).

While accurate, our current model is of course not ``the final word''
on modeling pentesting with POMDPs. As already mentioned, we currently
do not explicitly model the detrimental side-effects exploits may
have, i.e., the cases where they are detected (spawning a reaction of
the network defense) or where they crash a
machine/application. Another important aspect that could be modeled in
the POMDP framework is that machines are not independent. Knowing the
configuration of some computers in the network provides information
about the configuration of other computers in the same network. This
can be modeled in terms of the probability distribution given in the
initial belief. An interesting question for future research then is
how to generate these dependencies---and thus the initial belief---in
a realistic way. Answering this question could go hand in hand with
more realistically simulating the effect of the ``time delay'' in
pentesting. Both could potentially be adressed by learning appropriate
graphical models \cite{KolFri09}, based on up-to-date real-world
statistics.

To close the paper, it must be admitted that, in general, ``pentesting
$\neq$ POMDP solving'', by contrast to our paper title (hence the
question mark). Computer security is always evolving, so that the
probability of meeting certain computer configurations changes with
time. An ideal attacker should continuously learn the probability
distributions describing the network and computer configurations it
can encounter. This kind of learning can be done outside the POMDP
model, but there may be better solutions doing it more
natively. Furthermore, if the administrator of a target network reacts
to an attack, running specific counter-attacks, then the problem turns
into an adversarial game.

\bibliographystyle{aaai}

\bibliography{planning}

\end{document}